\documentclass[review]{elsarticle}

\usepackage{multirow}
\usepackage{longtable}
\usepackage{graphicx}
\usepackage{amsmath}
\usepackage{diagbox}
\usepackage[english]{babel}
\usepackage{graphicx}
\usepackage{graphicx}
\usepackage{epsfig}
\usepackage{epstopdf}
\usepackage{algorithmic}
\usepackage[linesnumbered,boxed]{algorithm2e}
\usepackage{amsmath}
\usepackage{diagbox}
\usepackage[english]{babel}
\usepackage{graphicx}
\usepackage{graphicx}
\usepackage{epsfig}
\usepackage{epstopdf}
\usepackage{amsthm}
\usepackage{mathrsfs}
\usepackage{amssymb}
\journal{Journal of \LaTeX\ Templates}
\begin{document}
\begin{frontmatter}
	\title{Weighted parallel SGD for distributed unbalanced-workload training system}

%
%
%
%

\author[mymainaddress,mysecondaryaddress]{Cheng Daning}
\ead{chengdaning@ict.ac.cn}

\author[mymainaddress]{Li Shigang\corref{mycorrespondingauthor}}
\ead{shigangli.cs@gmail.com}

\author[mymainaddress]{Zhang Yunquan}
\ead{zyq@ict.ac.cn}
\address[mymainaddress]{SKL of Computer Architecture, Institute of Computing Technology, Chinese Academy of Sciences, China}
\address[mysecondaryaddress]{University of Chinese Academy of Sciences}
\cortext[mycorrespondingauthor]{Corresponding author}

	

	\begin{abstract}
		Stochastic gradient descent (SGD) is a popular stochastic optimization method in machine learning. Traditional parallel SGD algorithms, e.g., SimuParallel SGD \cite{Zinkevich2010Parallelized}, often require all nodes to have the same performance or to consume equal quantities of data. However, these requirements are difficult to satisfy when the parallel SGD algorithms run in a heterogeneous computing environment; low-performance nodes will exert a negative influence on the final result. In this paper, we propose an algorithm called weighted parallel SGD (WP-SGD). WP-SGD combines weighted model parameters from different nodes in the system to produce the final output. WP-SGD makes use of the reduction in standard deviation to compensate for the loss from the inconsistency in performance of nodes in the cluster, which means that WP-SGD does not require that all nodes consume equal quantities of data. We also analyze the theoretical feasibility of running two other parallel SGD algorithms combined with WP-SGD in a heterogeneous environment. The experimental results show that WP-SGD significantly outperforms the traditional parallel SGD algorithms on distributed training systems with an unbalanced workload.
	\end{abstract}
	
\end{frontmatter}

\section{Introduction}
The training process in machine learning can essentially be treated as the solving of the stochastic optimization problem. The objective functions are the mathematical expectation of loss functions, which contain a random variable. The random variables satisfy a known distribution. The machine learning training process can be formalized as
\begin{align}
\min E[g( X,w)](X \sim \mbox{certain distribution } D) \notag\\
=\min \int_{\Omega}g( x,w)\mbox{$density$}(x)\Delta x
\end{align}
where $g(\cdot)$ is the loss function, $w$ is the variables, $X$ is the random variable, and $density(\cdot)$ is the probability density function of the distribution $D$.

Because some distributions cannot be presented in the form of a formula, we use the frequency to approximate the product of probability density $density(x)$ and $\Delta x$, as a frequency histogram can roughly estimate the curve of a probability density function. Thus, for a dataset, the above formula can be written in the following form:
\begin{align}
\min E[g( X ,w)](X \sim \mbox{certain distribution } D) \notag \\
\approx \min \dfrac{1}{m} \sum_{i = 1}^{m} g(  x^i,w)
\end{align}
where $m$ is the number of samples in the dataset, and $x^i$ is the $i$th sample value.

Stochastic gradient descent (SGD) is designed for the following minimization problem:
\begin{equation}
\min c(w)=\frac{1}{m}\sum\limits_{i=1}^{m}{{{c}^{i}}(w })	
\end{equation}
where $m$ is the number of samples in the dataset, and $c^i : \ell_2    \mapsto [ 0,\infty ]$ is a convex loss function indexed by $i$ with the model parameters $w \in {\mathbb{{R}}^{d}}$. Normally, in the case of regularized risk minimization, $c^i(w)$ is represented by the following formula:
\begin{equation}
{{c}^{i}}(w )=\frac{\lambda }{2}{{\left\| w  \right\|}^{2}}+L({{x}^{i}},{{y}^{i}},w \cdot {{x}^{i}})	
\end{equation}
where $L(\cdot)$ is a convex function in $w \cdot x$. It is of note that in the analysis and proof, we treat model parameters, i.e., $w$, as the random variable during the training process.

When $L(x^i,y^i,w \cdot x^i)$ is not a strong convex function, for example a hinge loss, the regularized term would usually guarantee the strong convexity for $c^i(w)$.

The  iteration step for sequential SGD is
\begin{equation}
{{w}_{n}}={{w}_{n-1}}-\eta {{\partial }_{w}}{{c}^{i}}({{w}_{n-1}})	
\end{equation}

Because of its ability to solve machine learning training problems, its small memory footprint, and its robustness against noise, SGD is currently one of the most popular topics \cite{Bottou2007The,Shalev2008SVM,Nemirovski2009Robust,Nesterov2009Primal,Dean2012Large,Dekel2012Optimal,Duchi2010Adaptive}.

As SGD was increasingly run in parallel computing environments \cite{abadi2016tensorflow,Jia2014Caffe}, parallel SGD algorithms were developed \cite{Zinkevich2010Parallelized,Feng2011HOGWILD}. However, heterogeneous parallel computing devices, such as GPUs and CPUs or different types of CPU, have different performance. The cluster may contain nodes having different computing performance. At the same time, parallel SGD algorithms suffer from performance inconsistency among the nodes \cite{Feng2011HOGWILD}. Therefore, it is necessary to tolerate a higher error rate or to use more time when running parallel SGD algorithms on an unbalanced-workload system.
\begin{algorithm}[h]
	\label{algorithm WP-SGD}
	\caption{WP-SGD}
	\KwIn{ Examples $\{{{c}^{1}},\dots,{{c}^{m}}\}$, learning rate $\eta $, nodes $k$\;}
	\KwOut{$v$}
	Randomly partition the examples\;
	\For{ all $i\in \{1,\dots,k\}$ parallel}
	{
		Randomly shuffle the data on machine $i$\;
		Initialize ${{w}_{i,0}}$ = 0\;
		Define the fastest nodes consuming $t$ samples\;
		Define the delay between the fastest node and the $i$th node as $T_i$\;
		\For {all $n\in \{1,\dots,t-T_i\}$}
		{
			Get the $n$th example on the $i$th node, ${{c}^{i,n}}$\;
			${{w}_{i,n}}={{w}_{i,n-1}}-\eta {{\partial }_{w}}{{c}^{i}}({{w}_{i,n-1}})$\;
		}
	}
	Aggregate from all nodes $v=\sum\limits_{i=1}^{k}{weight_{1-\eta \lambda,i}\cdot {{w}_{i,t}}}$\;
	Return $v$\;
\end{algorithm}

\begin{figure}[h]
	\label{working partten WP-SGD}
	\centering
	\includegraphics[width=6cm, height = 3.3cm]{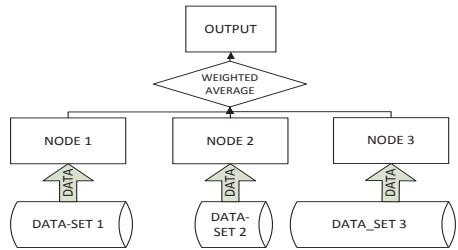}
	\caption{Working pattern of WP-SGD when the quantities of data differ}
\end{figure}

In this paper, we propose the following weighted parallel SGD (WP-SGD) for a distributed training system with an unbalanced workload. WP-SGD is given as Algorithm 1. WP-SGD adjusts the weights of model parameters from each node according to the quantity of data consumed by that node. The working pattern of WP-SGD is illustrated in Figure 1.

WP-SGD is based on SimuParallel SGD \cite{Zinkevich2010Parallelized}, which is shown as Algorithm 2. The working pattern of SimuParallel SGD is illustrated in Figure 2.
\begin{figure}[h]
	\label{working partten simuparallel SGD}
	\centering
	\includegraphics[width=6cm, height = 3.3cm]{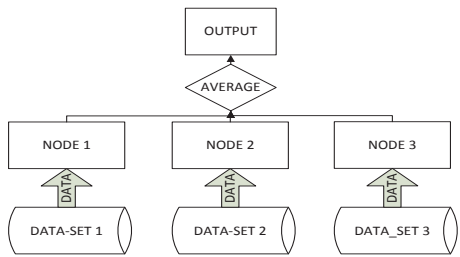}
	\caption{Working pattern of SimuParallel SGD with equal quantities of data}
\end{figure}

\begin{algorithm}[h]
	\label{algorithm SimuParalSGD}
	\caption{SimuParallel SGD}
	\KwIn{ Examples $\{{{c}^{1}},\dots,{{c}^{m}}\}$, learning rate $\eta $, nodes $k$\;}
	\KwOut{$v$}
	Randomly partition the examples\;
	\For{ all $i\in \{1,\dots,k\}$ parallel}
	{
		Randomly shuffle the data on machine $i$\;
		Initialize ${{w}_{i,0}}$ = 0\;
		All nodes consume $t$ samples\;
		\For {all $n\in \{1,\dots,t\}$}
		{
			Get the $n$th example on the $i$th node, ${{c}^{i,n}}$\;
			${{w}_{i,n}}={{w}_{i,n-1}}-\eta {{\partial }_{w}}{{c}^{i}}({{w}_{i,n-1}})$\;
		}
	}
	Aggregate from all nodes $v=\sum\limits_{i=1}^{k}{{\frac{1}{k}}\cdot {{w}_{i,t}}}$\;
	Return $v$\;
\end{algorithm}

In WP-SGD, when $L(\cdot)$ is not a strong convex function, we define $weight_{1-\lambda \eta,i}$ as follows:
\begin{equation}
weight_{1-\lambda \eta,i}^{{}}={{(1-\lambda \eta )}^{{{T}_{i}}}}/\sum\limits_{j=1}^{k}{{{(1-\lambda \eta )}^{{{T}_{j}}}}}
\end{equation}
where $(1-\lambda\eta)$ is the contracting map rate for the SGD framework.

The main bottleneck for SimuParallel SGD in the heterogeneous parallel computing environment is that we need to guarantee that all nodes have trained on equal quantities of data before we average them (Line 5 and Line 11, respectively, in Algorithm 2). This requirement leads to a degradation in performance on the heterogeneous cluster. WP-SGD uses a weighted average operation to break this bottleneck. WP-SGD does not require all nodes to be trained on equal quantities of data and incorporates the delay information into the weights (Line 5, Line 6, and Line 12 with Eq. 6), which allows WP-SGD to run efficiently in a heterogeneous parallel computing environment.

WP-SGD suggests that when the workload is unbalanced within the cluster and there is a delay between the fastest node and the $i$th node, the weight of the model parameters on the $i$th node should be decreased exponentially.

Under some conditions, the upper bound of the objective function value calculated by the output of WP-SGD will be less than the upper bound of the objective function value of sequential SGD in the fastest node. When the standard deviation of the distribution corresponds to the fixed point of the model parameters, the standard deviations of the dataset are large enough, and $L(x^i,y^i,w \cdot x^i)$ is not a strong convex function, the above conditions are

\begin{equation}
2\sum\limits_{i=1}^{k}{{{(1-\eta \lambda )}^{{{T}_{i}}}}}>\sqrt{k}+k
\end{equation}

Furthermore, for the case that the contracting map rate of $c(w)$ in SGD is much smaller than the contracting map rate of the SGD framework in view of the whole process, which is due to the fact that $L(x^i,y^i,w \cdot x^i)$ may be a strong convex function, we should choose a smaller contracting map rate (we denote it as $r$) to replace  $1 - \lambda  \eta$:
\begin{equation}
weight_{r,i}={{r}^{{{T}_{i}}}}/\sum\limits_{j=1}^{k}{{{r}^{ {{T}_{j}}}}}					
\end{equation}
Additionally, under the following limitation, the output from WP-SGD will outperform the output from the fastest nodes:
\begin{equation}
2\sum\limits_{i=1}^{k}{{{r}^{{{T}_{i}}}}}>\sqrt{k}+k
\end{equation}
The value of $r$ is determined via experience, data fitting, or analysis of the training data and $L(\cdot)$.

A numerical experiment on data from KDD Cup 2010 \cite{Yu2010Feature} shows that the final output of WP-SGD with an unbalanced workload can be nearly equivalent to the output from a system with a perfectly balanced workload. In a workload-unbalanced environment, WP-SGD uses less time than workload-balanced SGD. To clearly show the gap between different algorithms, we also conducted experiments using analog data; these experiments show that WP-SGD is able to handle cases in which there is unbalanced workload among the nodes.

The key contributions of this paper are as follows:

1. We propose a novel parallel SGD algorithm, WP-SGD, for distributed training system with unbalanced workloads.

2. We theoretically prove that WP-SGD can tolerate a large delay between different nodes. WP-SGD suggests that when there is an increase in the delay between the fastest node and the $i$th node, the weight of the model parameters for the $i$th node should be decreased exponentially.

3. We provide the results of experiments which we conducted using analog data and real-world data to demonstrate the advantages of WP-SGD on computing environment with unbalanced workloads.

In the next section, we present a basic view of traditional parallel SGD algorithms. In Section III, we demonstrate the basic theory of SGD and provide the proof for WP-SGD. In Section IV, we theoretically offer some complementary technologies based on WP-SGD. In Section V, we present the results of the numerical experiments.

\section{Related work}
SGD dates back to early work by Robbins and Monro et al. \cite{Nemirovski2009Robust,Bottou2010Large}. In recent years, combined with the GPU \cite{abadi2016tensorflow,Jia2014Caffe}, parallel SGD algorithms have become one of the most powerful weapon for solving machine learning training problems \cite{Dekel2012Optimal,Duchi2010Distributed,Langford2009Slow}. Parallel SGD algorithms can be roughly classified into two categories, which we call delay SGD algorithms and bucket SGD algorithms.

Delay SGD algorithms first appeared in Langford et al.'s work \cite{Langford2009Slow}. In a delay SGD algorithm, current model parameters add the gradient of older model parameters in $\tau $ iterations ($\tau$ is a random number where $\tau <M$, in which $M$ is a constant). The iteration step for delay SGD algorithms is
\begin{equation}
{{w}_{n}}={{w}_{n-1}}-\eta {{\partial }_{w}}{{c}^{i}}({{w}_{n-\tau}})	
\end{equation}
In the Hogwild! algorithm \cite{Feng2011HOGWILD}, under some restrictions, parallel SGD can be implemented in a lock-free style, which is robust to noise \cite{chaturapruek2015asynchronous}. However, these methods lead to the consequence that the convergence speed will be decreased by o(${{\tau }^{2}}$). To ensure the delay is limited, communication overhead is unavoidable, which hurts performance. The trade-off in delay SGD is between delay, degree of parallelism, and system efficiency:

\textbf{1.} Low-lag SGD algorithms use fewer iteration steps to reach the minimum of the objective function. However, these algorithms limit the number of workers and require a barrier, which is a burden when engineering the system.

\textbf{2.} Lock-free method is efficient for engineering the system, but the convergence speed, which depends on the maximum lag, i.e. $\tau$ in Eq.10, is slow.

\textbf{3.} The lower limit of the delay is the maximum number of workers the system can have.

From the point of view of engineering implementation, the implementation of delay SGD algorithms is accomplished with a parameter server. Popular parameter server frameworks include ps-lite in MXNet \cite{Chen2015MXNet}, TensorFlow \cite{abadi2016tensorflow}, and Petuum \cite{Xing2013Petuum}. A method that constricts the delay was offered by Ho et al. \cite{Ho2013More}. However, if the workers in the parameter server have different performance, $\tau$ is increased, causing convergence speed to be reduced.

Delay SGD algorithms can be considered as an accelerated version of sequential SGD. Bucket SGD algorithms accelerate SGD via the averaging of model parameters. Zinkevich et al. \cite{Zinkevich2010Parallelized} proposed SimuParallel SGD, which has almost no communication overhead. Y. Zhang et al. \cite{Zhang2012Communication} gave a insightful analysis and proof for this parallel algorithm. However, these methods do not take  into account the heterogeneous computing environment. J. Zhang et al. \cite{Zhang2012Communication} also point out the invalidity of SimuParallel SGD. In fact, the effect of a bucket SGD depends primarily on how large the model parameters' relative standard deviation is, which means it is a trade-off between the parallelism and the applicability for dataset.

From the point of view of engineering implementation, Bucket SGD algorithms can be implemented in a MapReduce manner \cite{Dean2004MapReduce}. Thus, most of them are running on platforms like  Hadoop \cite{White2010Hadoop}. If the nodes in the cluster have different performance, the slowest node is the performance bottleneck.

Along with parallel SGD algorithms, many other kinds of numerical optimization algorithms have been proposed, such as PASSCoDe \cite{Cho2015PASSCoDe} and CoCoA \cite{Jaggi2014Communication}. They share many new features, such as fast convergence speed in the end of training phase. Most of them are formulated from the dual coordinate descent (ascent) perspective, and hence can only be used for problems whose dual function can be computed. Moreover, traditional SGD still plays an important role in those algorithms.

These parallel SGD algorithms have various superb features. However, all of them lack robustness against an unbalanced workload.
	
	\section{Proof and analysis}
	\subsection{Notation and definitions}
	We collect our common notations and definitions in this subsection.
	
	\textbf{Definition 1} (Lipschitz continuity) A function $f$:$\mathcal{X}  \mapsto \mathbb{R}$ is Lipschitz continuous with constant $C$ with respect to a distance $d$ if $| f(x) - f(y)| \le Cd(x,y)$ for all $x,y \in \mathcal{X}$.
	
	\textbf{Definition 2} (Lipschitz seminorm) Luxburg and Bousquet \cite{Luxburg2003Distance} introduced a seminorm. With minor modification, we use
	\begin{align}
	&{{\left\| f \right\|}_{\mathrm{Lip}}} \notag\\
	&:= inf\{C \mbox{ where } |f(x) - f(y)| \le Cd(x,y) \mbox{ for all } x,y \in \mathcal{X} \}
	\end{align}
	That is, $\left\| f \right\|_{\mathrm{Lip}}$ is the smallest constant for which Lipschitz continuity holds.
	
	In the following, we let  ${{\left\| L(x,y,y') \right\|}_{\mathrm{Lip}}} \le G$ as a function of $y'$ for all occurring data $(x,y) \in \mathcal{X} \times \mathcal{Y}$ and for all values of $w$  within a suitably chosen (often compact) domain. $G$ is a constant.
	
	\textbf{Definition 3} (relative standard deviation of $X$ with respect to $a$)
	\begin{equation}
	\sigma _{X}^{a}=\sqrt{E{{(X-a)}^{2}}}					
	\end{equation}
	As we can see, $\sigma _{X}=\sigma _{X}^{{{\mu }_{X}}}$, where ${\mu}_{X}$ is the mean of $X$.
	
	Table I shows the notations used in this paper and the corresponding definitions.
		\begin{longtable}{|c|m{0.5\columnwidth}|}
			\hline
			Notation	&Definition\\\hline\hline
			\endhead
			$t$		&the number of samples consumed by the fastest nodes\\
			\hline
			$T_i$	&the delay between the fastest nodes and the $i$th node\\
			\hline
			$x^j$	&the $j$th sample value\\
			\hline
			$y^j$	&the label for the $j$th sample\\
			\hline
			$\lambda$	&the parameter for the regularization term. For some loss functions, such as hinge loss, it guarantees strong convexity.\\
			\hline
			$\eta$	&step length or learning rate for SGD\\
			\hline
			$w$		&variables for function and for machine learning. It is the model parameters.\\
			\hline
			$X$		&the random variable\\
			\hline
			$m$		&the number of samples in the dataset\\
			\hline
			$c(\cdot)$	&loss function \\
			\hline
			$weight_{r,i}$	&in WP-SGD, the weight for the $i$th node on contracting map rate $r$\\
			\hline
			$L(x^j,y^j,w\cdot x^j)$		&the loss function without a regularization term\\
			\hline
			$D$		&the distribution for the random variables\\
			\hline
			$k$		&the total number of nodes in a cluster\\
			\hline
			$r$		&the contracting map rate for $c(w)$ in SGD\\
			\hline
			$\tau$	&in delay SGD, the delay between the current model parameters and the older model parameters\\
			\hline
			$M$  & maximum number of $\tau$\\
			\hline
			$D^*_\eta$	&the distribution of the unique fixed point in SimuParallel SGD and WP-SGD, with learning rate $\eta$\\
			\hline
			$D^t_\eta$ &the distribution of the stochastic gradient descent update after $t$ updates, with learning rate $\eta$.\\
			\hline
			$W^{i,t-T_i}$	&the output of the $i$th node after $t-T_i$ iterations.\\
			\hline
			$W^{\#,t}$  & the output of WP-SGD, where the fastest node trained on $t$ samples\\
			\hline
			$D^{\#,t}_\eta$	&the distribution of $W^{\#,t}$\\
			\hline
			$Wasserstein_z(X,Y)$	&Wasserstein distance between two distributions $X,Y$\\
			\hline
			$span$	&in more average operation SimuParallel SGD and WP-SGD, which are offered at Section IV, the span between two average operations from the view of the fastest nodes\\
			\hline
			$v$ &the final output of an algorithm\\
			\hline
			\caption{Notations and definitions}
		\end{longtable}
	
\subsection{Introduction to SGD theory}
Theorem 1, Theorem 2, Theorem 3, and Lemma 1 are key theorems we will use. All four theorems are proved by Zinkevich et al. \cite{Zinkevich2010Parallelized}.

\textbf{Theorem 1} Given a cost function $c$ that ${{\left\| c \right\|}_{\mathrm{Lip}}}$ and ${{\left\| \nabla c \right\|}_{\mathrm{Lip}}}$ are bounded, and a distribution $D$ such that ${{\sigma }_{D}}$ is bounded, then for any point $p$
\begin{align}
&{{E}_{p \in D}}[c(p)]-\underset{w}{\mathop{\min }}\,c(w)\le \notag\\
&\sigma _{D}^{p}\sqrt{2{{\left\| \nabla c \right\|}_{\mathrm{Lip}}}(c(p)-\underset{w}{\mathop{\min }}\,(w))} \notag\\
&+({{\left\| \nabla c \right\|}_{\mathrm{Lip}}}{{(\sigma _{D}^{p})}^{2}}/2)+(c(p)-\underset{w}{\mathop{\min }}\,c(w))
\end{align}

Theorem 1 highlights the relationship between the distribution of model parameters and $\underset{w\in {\mathbb{R}^{d}}}{\mathop{\min }}\,c(w)$, which is the expected result of SGD when  $p$ is equal to $w$.

\textbf{Theorem 2}
\begin{equation}
c({{E}_{w\in D_{\eta }^{*}}}[w])-{{\min }_{w\in {\mathbb{R}^{d}}}}c(w)\le 2\eta G^2
\end{equation}
where $D_{\eta }^{*}$ is the distribution of a unique fixed point in SimuParallel SGD.
This theorem provides an idea of the bound on the third part of Theorem 1.

\textbf{Lemma 1}
\begin{equation}
\sigma _{X}^{c}\le \sigma _{X}^{c'}+d(c,c')
\end{equation}
where $d(\cdot ,\cdot )$ is the Euclidean distance.

\textbf{Theorem 3} If $D_{\eta }^{t}$ is the distribution of the stochastic gradient descent update after $t$ iterations, then
\begin{equation}
d({{\eta }_{D_{\eta }^{t},}}{{\eta }_{D_{\eta }^{*}}})\le \frac{G}{\lambda }{{(1-\eta \lambda )}^{t}}
\end{equation}
\begin{equation}
\sigma _{D_{\eta }^{t}}^{{}}\le \frac{2\sqrt{\eta }G}{\sqrt{\lambda }}+\frac{G}{\lambda }{{(1-\eta \lambda )}^{t}}		
\end{equation}

The above theorems describe how and why SGD can converge to a minimum.
The difference between the value of $c(\cdot)$ using the output $w$ from SGD and the minimum of $c(\cdot)$ is controlled by three factors:

\textbf{(1)} The difference between the expectation of the current distribution of model parameters and the expectation of $D_{\eta }^{*}$

\textbf{(2)} The standard deviation of the distribution of the current model parameters, which is $\sigma _{D_{\eta }^{t}}^{{}}$

\textbf{(3)} The difference between the expected value of $c(w)$ when $w$ satisfies distribution $D_{\eta }^{*}$ and the minimum value of $c(\cdot)$

For the sequential SGD, carrying out the algorithm would reduce the first part and the second part. The third part is controlled by $\eta$ and $L(\cdot)$.

For SimuParallel SGD, the first part and the third part are the same for different nodes. However, $\sigma _{D_{\eta }^{t}}^{{}}$ can benefit from the averaging operation. SimuParallel SGD uses the gain in the standard deviation to reduce the number of iteration steps needed to reduce the first and second parts. In other words, SimuParallel SGD accelerates SGD.

\subsection{Analysis of WP-SGD}
The concept of WP-SGD has two main aspects:

1. Our proposed weight is to compensate for the main loss from the delay between the different nodes. The main loss from the delay is controlled by the exponential term $(1-\lambda\eta)^t$.

2. Under the condition that the gain from the standard deviation's reduction is greater than the loss in the mean's weighted average from the perspective of the fastest node, the WP-SGD output will outperform the fastest node.

All of the following lemmas, corollaries, and theorems are our contributions.

We focus on the first aspect at the beginning:
Corollary 1 and Lemma 3 show how the mean and standard deviation will change by using WP-SGD. Their sum is the upper bound of the relative standard deviation which is shown in Lemma 1.

Lemma 2 is used in the proof of  Corollary 1.

\textbf{Lemma 2}  Suppose that ${{X}^{1}} \dots {{X}^{k}},B $ are independent distributed random variables over ${\mathbb{R}^{d}}$. Then if $A=\sum\limits_{i=1}^{k}{weigh{{t}_{i}}\cdot {{X}^{i}}}$ and $1=\sum\limits_{i=1}^{k}{weigh{{t}_{i}}}$, it is the case that
\begin{equation}
d({{\mu }_{A}},{{\mu }_{B}})\le \sum\limits_{i=1}^{k}{weigh{{t}_{i}}\cdot d({{\mu }_{{{X}^{i}}}},{{\mu }_{B}})} \notag
\end{equation}

\textbf{Corollary 1} The fastest node consumes $t$ data samples. $D_{\eta }^{t-{{T}_{i}}}$ is the distribution of model parameters updated after $t-{{T}_{i}}$ iterations in node $i$, and $D_{\eta }^{\#,t}$ is the distribution of the  stochastic gradient descent update in WP-SGD.
\begin{equation}
d({{\eta }_{D_{\eta }^{\#,t}}},{{\eta }_{D_{\eta }^{*}}})\le \frac{Gk{{(1-\eta \lambda )}^{t}}}{\lambda \sum\limits_{i=1}^{k}{{{(1-\eta \lambda )}^{{{T}_{i}}}}}}
\end{equation}

\textbf{Lemma 3}  ${{W}^{i,t-{{T}_{i}}}}$ is the output of node $i$. Then, if
\begin{equation}
{{W}^{\#,t}}=\sum\limits_{i=1}^{k}{weigh{t}_{1-\eta \lambda,i}\cdot {{W}^{i,t-{{T}_{i}}}}}
\end{equation}
then the distribution of ${W}^{\#,t}$ is $D^{\#,t}_{\eta}$. It is the case that
\begin{equation}
\sigma _{{{D_\eta}^{\#,t}}}\le \frac{\sqrt{k}}{(\sum\limits_{i=1}^{k}{{{(1-\eta \lambda )}^{{{T}_{i}}}})}}(\frac{2G\sqrt{\eta }}{\sqrt{\lambda }}+\frac{G}{\lambda }{{(1-\eta \lambda )}^{t}})
\end{equation}

Combining Lemma 1, Theorem 1, Corollary 1 whose proof uses Lemma 2, and Lemma 3, we have the following:

\textbf{Theorem 4} Given a cost function $c$ such that ${{\left\| c \right\|}_{\mathrm{Lip}}}$ and ${{\left\| \nabla c \right\|}_{\mathrm{Lip}}}$ are bounded, the bound of WP-SGD is
\begin{align}
& {{E}_{w\in D}}[c(w)]-\underset{w}{\mathop{\min }}\,c(w) \notag \\
&\le ((\frac{Gk{{(1-\eta \lambda )}^{t}}}{\lambda(\sum\limits_{j=1}^{k}{{{(1-\eta \lambda )}^{{{T}_{j}}}}})}+\frac{\sqrt{k}}{\sum\limits_{j=1}^{k}{{{(1-\eta\lambda)}^{{{T}_{j}}}}}}(\frac{2G\sqrt{\eta }}{\sqrt{\lambda }} \notag
\\
& +\frac{G}{\lambda }{{(1-\eta \lambda )}^{t}}))\sqrt{2{{\left\|\nabla c \right\|}_{\mathrm{Lip}}}}+  \sqrt{c(v)-\underset{w}{\mathop{\min }}\,c(w)}{{)}^{2}}
\end{align}

Next, we discuss the second aspect.

It is apparent that there is no guarantee that the output of WP-SGD will be better than the output from the fastest nodes, because from the viewpoint of the best-performing node, the weighted average will damage its gain from contraction of the mean value term. Here, we offer the Corollary 2 that defines the conditions under which the output from the fastest nodes will benefit from the normal-performance nodes. In the following, $Wasserstein_z(X,Y)$ is the Wasserstein distance between two distributions $X,Y$, and the fastest nodes consume $t$ data samples in an unbalanced-workload system.

\textbf{Corollary 2}  For WP-SGD, when
\begin{align}
&\dfrac{\sum\limits_{i=1}^{k}{{{(1-\eta \lambda )}^{{{T}_{i}}}}}- \sqrt{k}}{k-\sum\limits_{i=1}^{k}{{{(1-\eta \lambda )}^{{{T}_{i}}}}}} \notag \\
&> \dfrac{(1-\lambda\eta)^t Wasserstein_1({{D_{\eta }^{\#,1}}},{{D_{\eta }^{*}}}) }{(1-\lambda\eta)^t \cdot Wasserstein_2({{D_{\eta }^{\#,1}}},{{D_{\eta }^{*}}}) + \sigma_{D_\eta^*}}
\end{align}
the upper bound of the objective function value of WP-SGD is closer to the minimum than is the upper bound of the objective function value of sequential SGD on the fastest nodes.

$Wasserstein_z({{D_{\eta }^{\#,1}}},{{D_{\eta }^{*}}})$ is not a prior value. However, Corollary 2 still eliminates the dataset whose $\sigma_{D_\eta^*}$ and  $\sigma_{D_\eta^{\#,1}}$ are small. $\sigma_{D_\eta^{\#,1}}$ is the standard deviation of the dataset. The standard deviation of the dataset will influence the values of $ Wasserstein_1({{D_{\eta }^{\#,1}}},{{D_{\eta }^{*}}})$ and  $ Wasserstein_2({{D_{\eta }^{\#,1}}},{{D_{\eta }^{*}}})$. In an extreme example, when all samples in the dataset are the same, i.e., SGD degenerates into Gradient Descent, i.e. GD, WP-SGD would be invalid, and this is also the case with SimuParallel SGD.

Most of the time, the standard deviations of real-world datasets are usually large enough. In the case where the $\sigma_{D_\eta^{\#,1}}$ and  $\sigma_{D_\eta^*}$  are large enough, under Corollary 3, WP-SGD would be better than the sequential SGD.

\textbf{Corollary 3} For WP-SGD, on a dataset having a large standard deviation and a large standard deviation of the fixed point, when
\begin{equation}
2\sum\limits_{i=1}^{k}{{{(1-\eta \lambda )}^{{{T}_{i}}}}}>\sqrt{k}+k
\end{equation}
the upper bound of the objective function value of WP-SGD is closer to the minimum than is the upper bound of the objective function value of sequential SGD on the fastest nodes.

Corollary 3 suggests that WP-SGD can tolerate sufficient delay. As we can see, the robustness of whole system will be stronger as the scale of the cluster increases.

\subsection{Analysis and redesign: weight for the dataset and $c(\cdot)$ whose contracting map rate is small}
When considering the equivalent condition of inequalities, it is obvious that $(1-\eta \lambda )$ is the best contracting map rate choice in the overall process only when $L(\cdot)$ is very close to being a linear function (the proof of Lemma 3 in Zinkevich et al.'s work \cite{Zinkevich2010Parallelized}). This requirement means that $L(\cdot)$ is not a strong convex function.

In fact, $(1-\eta \lambda )$ is the upper bound of the contracting map rate for every iteration. Yet the contracting map rate varies during the iteration process for every iteration, though it is always less than $(1-\eta \lambda )$. When the loss function's second derivative is larger, or during the process, many of the samples' directions are parallel to the current model parameters' direction, the contracting map rate will be smaller. Therefore, from the standpoint of the overall iteration process rather than that of a single iteration, we should redesign a smaller contracting map rate to replace $(1-\eta \lambda )$. We denote this new contracting map rate by $r$. Usually, $r$ should be a smaller number when the direction of processing samples is closer to the direction of the current model parameters, i.e., $w_n$, and the second derivative of $L(\cdot)$ is larger.

We can determine the value of the new contracting map parameter via experience, data fitting, or analysis of training data and $L(\cdot)$, as in Figure 3.

\begin{figure}[!ht]
	\label{An example of using contracting map rate $r$ fitting the real contracting process. In this example, the objective function value decrease from 12000 to zero in 500000 iterations}
	\centering
	\includegraphics[width=9.5cm, height = 4.7cm]{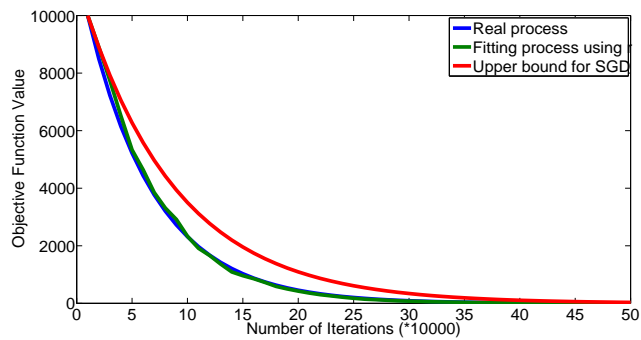}
	\caption{A example of using contracting map rate $r$ for fitting the actual contracting process. In this example, the objective function value decrease from 12000 to zero in 500000 iterations}
\end{figure}

As we ascertain a value for the new contracting map rate $r$, we rewrite $weight$, Theorem 4 and Corollary 3 as follows:
\begin{equation}
weight_{r,i}^{{}}={{r}^{{{T}_{i}}}}/\sum\limits_{j=1}^{k}{{{r}^{ {{T}_{j}}}}}	\notag				
\end{equation}
\textbf{Theorem 5} (incorporating $r$ into Theorem 4) Given a cost function $c$ such that ${{\left\| c \right\|}_{\mathrm{Lip}}}$ and ${{\left\| \nabla c \right\|}_{\mathrm{Lip}}}$ are bounded, and in view of the overall process, the contracting map rate is $r$, and the bound of WP-SGD is
\begin{align}
& {{E}_{w\in D}}[c(w)]-\underset{w}{\mathop{\min }}\,c(w)\le ((\frac{Gk{{r}^{t}}}{\lambda (\sum\limits_{j=1}^{k}{{{(r)}^{{{T}_{j}}}}})}+\frac{\sqrt{k}}{\sum\limits_{j=1}^{k}{{{(r )}^{{{T}_{j}}}}}}(\frac{2G\sqrt{\eta }}{\sqrt{\lambda }} \notag \\
&+\frac{G}{\lambda }{{(r )}^{t}}))\sqrt{2{{\left\| \nabla c \right\|}_{\mathrm{Lip}}}}  +\sqrt{c(v)-\underset{w}{\mathop{\min }}\,c(w)}{{)}^{2}}
\end{align}

\textbf{Corollary 4} (incorporating $r$ into Corollary 3) Given that WP-SGD runs on a dataset having a large standard deviation and a large standard deviation of the fixed point, and in view of the overall process, the contracting map rate of $c(\cdot)$ is $r$, and when
\begin{equation}
2\sum\limits_{i=1}^{k}{{{r}^{{{T}_{i}}}}}>\sqrt{k}+k
\end{equation}
the upper bound of the objective function value of WP-SGD is closer to the minimum than is the upper bound of the objective function value of sequential SGD on the fastest nodes.

\section{Theoretical feasibility of efficiently running popular parallel SGD algorithms combined with WP-SGD in heterogeneous environments}

Current parallel SGD algorithms lack the feature of robustness in heterogeneous environments. However, they are characterized by a number of superb features such as the overlap between communication and computing (delay SGD) and fast convergence speed (bucket SGD). It is reasonable to consider combining WP-SGD with these algorithms in order to gain the benefits of their excellent features and the adaptability to unbalanced-workload environments. Considering the propose of this paper is introducing WP-SGD instead of developing other kinds of parallel SGD algorithm and the experiments are exhausting, we only theoretically offer methods via which we could combine current parallel SGD algorithms and WP-SGD, to show the theoretical feasibility of running current parallel SGD algorithms in a heterogeneous environment with help from WP-SGD.

\subsection{Combining WP-SGD with bucket SGD}
Although bucket SGD is not the most popular parallel SGD, the main idea of bucket SGD is reflected in the popular mini-batch style of SGD that averages the model parameters at each iteration \cite{Zhang2016Parallel}. However, averaging at each iteration operation is expensive, and the mini-batch is more vulnerable to performance differences. There is a compromise parallel SGD algorithm that averages model parameters at a fixed $span$ length. The number of $span$ is from the point of the best performance nodes. Here we offer theoretical analyses of this parallel algorithm and its theoretical performance in unbalanced-workload environments, based on the analyses of WP-SGD.

\textbf{Deduction 1} Given a cost function $c$ such that ${{\left\| c \right\|}_{\mathrm{Lip}}}$ and ${{\left\| \nabla c \right\|}_{\mathrm{Lip}}}$ are bounded, we average parameters every $span$ iterations for the fastest node in SimuParallel SGD. Then, the bound of the algorithm is
\begin{align}
&{{E}_{w\in D_{\eta }^{T,k}}}[c(w)]-\underset{w}{\mathop{\min }}\,c(w)\le \notag\\
& ((\frac{G{{(1-\eta \lambda )}^{t}}}{\lambda }+\frac{1}{{{(\sqrt{k})}^{t/span}}}(\frac{2G\sqrt{\eta }}{\sqrt{\lambda }} \notag \\
&+\frac{G}{\lambda }{{(1-\eta \lambda )}^{t}}))\sqrt{2{{\left\| \nabla c \right\|}_{\mathrm{Lip}}}} \notag\\
&+\sqrt{c(w){{|}_{w\in D_{\eta }^{D,k}}}-\underset{w}{\mathop{\min }}\,c(w)}{{)}^{2}}
\end{align}

\textbf{Deduction 2} Given a cost function $c$ such that ${{\left\| c \right\|}_{\mathrm{Lip}}}$ and ${{\left\| \nabla c \right\|}_{\mathrm{Lip}}}$ are bounded, we average parameters every $span$ iterations for the fastest node in WP-SGD. Then, the bound of the algorithm is
\begin{align}
& {{E}_{w\in D}}[c(w)]-\underset{w}{\mathop{\min }}\,c(w) \notag \\
&\le ((\frac{G{{(1-\eta \lambda )}^{t}}}{\lambda }\cdot {{(\frac{k}{(\sum\limits_{j=1}^{k}{{{(1-\eta \lambda )}^{{{T}_{j}}}}})})}^{t/span}}+ \notag \\
& {{(\frac{\sqrt{k}}{\sum\limits_{j=1}^{k}{{{(1-\eta \lambda )}^{{{T}_{j}}}}}})}^{t/span}}(\frac{2G\sqrt{\eta }}{\sqrt{\lambda }}+\frac{G}{\lambda }{{(1-\eta \lambda )}^{t}}))\sqrt{2{{\left\| \nabla c \right\|}_{\mathrm{Lip}}}} \notag \\
& +\sqrt{c(v)-\underset{w}{\mathop{\min }}\,c(w)}{{)}^{2}}
\end{align}

For all nodes with the same performance, the more average the operation, the closer the output model parameters will be to the function minimum. In this case, our consideration should be to balance the cost of operation and the gain from the ``better'' result. As is well known, not all training datasets' variances are large enough to get the expected effect. On an unbalanced-workload system, we should also guarantee that $(\frac{k}{(\sum\limits_{j=1}^{k}{{{(1-\eta \lambda )}^{{{T}_{j}}}}})})\cdot {{(1-\eta \lambda )}^{span}}<1$ to ensure overall that the training process is valid.

\subsection{Combining WP-SGD with delay SGD}
Because of the excellent adaptability on different kinds of datasets and the overlapping of the cost of communication and computing, delay SGD is widely used in machine learning frameworks such as MXNet \cite{Chen2015MXNet}, TensorFlow \cite{abadi2016tensorflow}, and Petuum \cite{Xing2013Petuum}. However, all of these algorithms are designed for a balanced-workload environment. In this section, we offer Algorithm 3, which combines WP-SGD and one kind of delay SGD to make delay SGD algorithms work efficiently in heterogeneous computing environments. Some intermediate variables are defined in the algorithm description. The working pattern of Algorithm 3 is illustrated in Figure 4.

\begin{algorithm}[!htbp]
	\caption{WP-SGD and delay SGD}
		\KwIn { Examples $\{{{c}^{1}},\dots,{{c}^{m}}\}$, learning rate $\eta $, nodes $k$\;}
		\KwOut{$v$}
		
		Randomly partition the examples\;
		\textbf{Phase 1:}\\
		\textbf{For Worker:}\\
		$pull$ the $w_{i,j}$ from the $i$th Server\;
		calculate ${{\partial }_{w}}{{c}_{i,j}}({{w}_{i,j}})$\;
		$push$ the ${{\partial }_{w}}{{c}_{i,j}}({{w}_{i,j}})$ to the Server\;
		
		\textbf{For the $i$th Server}\\
		
		Initialize ${{w}_{i,0}}= 0$\;
		\For{$j \in  (0 \dots Forever)$}{ 
			receive	 ${{\partial }_{w}}{{c}_{i,j-1-\tau}}({{w}_{i,j-1-\tau}})$ from the Worker\;
			Initialize $Flag = true$\;
			$Call$ function $Check(w_{j-1-\tau}\cdots w_{j-1},\lambda,\eta,x^j,Flag)$\; 
			\If{Flag}{
				${{w}_{i,j}} :={{w}_{i,j-1}}-\eta {{\partial }_{w}}{{c}_{i,j}}({{w}_{i,j-1-\tau}})$\;
				$Call$ function $Check(w_{j-2}\cdots w_{j},\lambda,\eta,x^j,Flag)$\; 
			}
			
			\If{!Flag}{
			abandon 	${{w}_{i,j}}$\;
			j = j -1\;
		}
	}
	\textbf{Phase 2:}\\
	Aggregate $v$ from all Servers $v=\sum\limits_{i=1}^{k}{weight_{r,i}\cdot {{w}_{i,j}}}$\;
	Return $v$\;
\end{algorithm}

\begin{algorithm}[!htbp]
	\caption{$Check$ function}
	\KwIn { model parameters $\{{{w}_{j-1-\tau}},\dots,{{w}_{j-1}}\}$, regularization parameter $\lambda$,learning rate $\eta $, sample $x^j$, Output $Flag$;}
	
		\For {all $ j_{tmp}\in \{j-1-\tau,\dots,j\}$}
		{
			$Length_{j_{tmp}} := x^{j} \cdot w_{j_{tmp}}$\;
			$Length_{j_{tmp}-1} := x^{j} \cdot w_{j_{tmp}-1}$\;
			$Length_{j_{tmp}-2} := x^{j} \cdot w_{j_{tmp}-2}$\;
			$\beta^{2} := x^{j} \cdot x^{j}$\;
			$Length_{j_{tmp}-1\perp} := w_{j_{tmp}-1} - Length_{j_{tmp}-1}/\sqrt{\beta}$\;
			$Length_{j_{tmp}-2\perp} := w_{j_{tmp}-2} - Length_{j_{tmp}-2}/\sqrt{\beta}$\;
			$c^{*} := \left\|                \dfrac{{\partial}L(y, \hat y)}{\partial \hat y}           \right\|$\;
			
			$rate := \sqrt[\tau]{\lambda\eta + c^{*}\eta\beta^{2} }$\; 			                                                                                                                                			 $Length_{min} :=\dfrac{Length_{j_{tmp}-1} - rate \cdot Length_{j_{tmp}-2}}{ 1 - rate }$\;
			
			\If {((${{Length}_{j_{tmp}}} \notin [Length_{min},Length_{j_{tmp}-1}]$ and ${{Length}_{j_{tmp}}} \notin [Length_{j_{tmp}-1},Length_{min}]$) or $\dfrac{Length_{j-2\perp}}{Length_{j-1\perp}}>1$)}
			{
				$Flag = false$\;
			}						
		}

\end{algorithm}

\begin{figure}[!ht]
	
	\centering
	\includegraphics[width=9cm, height = 5cm]{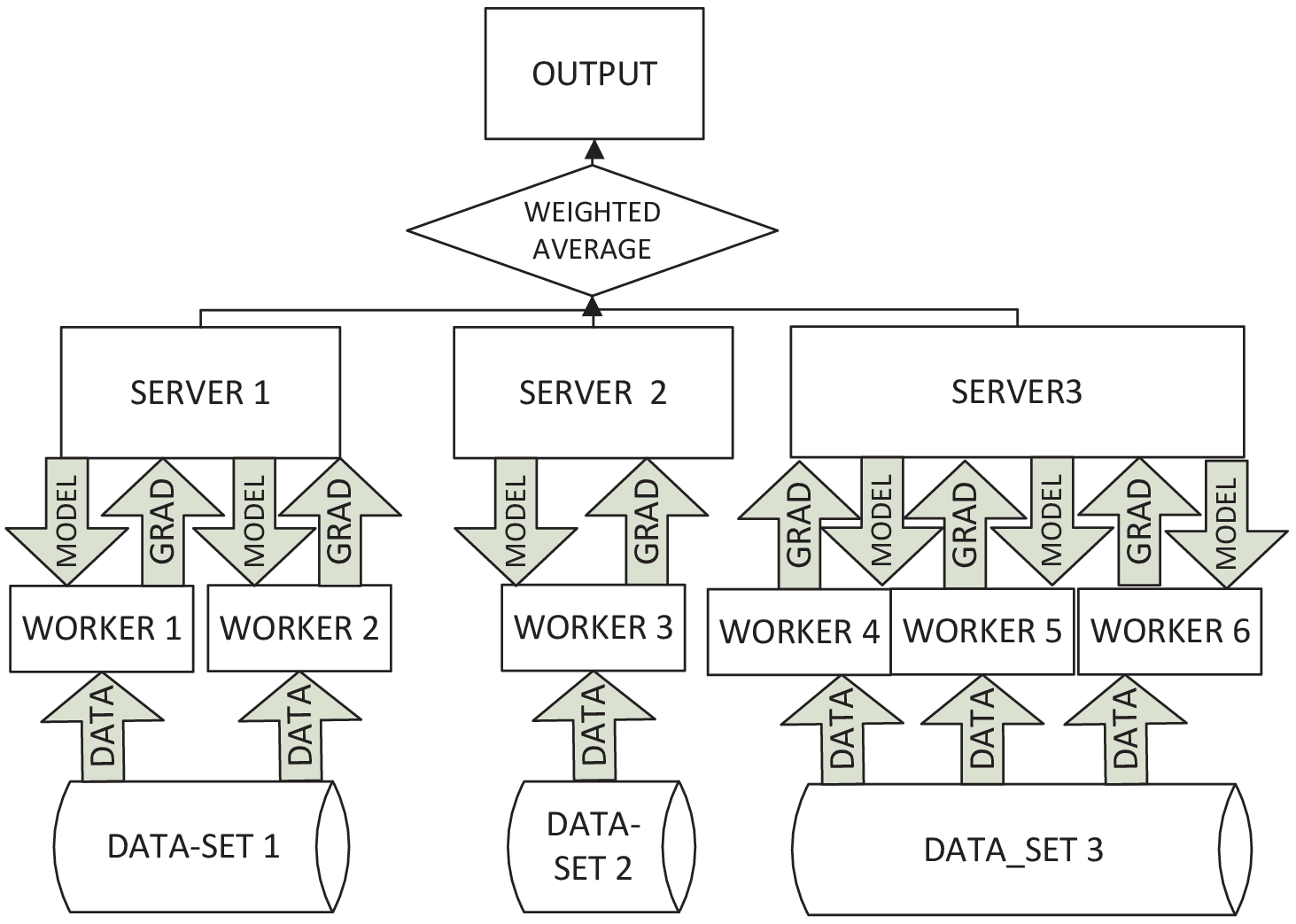}
	\caption{Working pattern of Algorithm 3 when the quantities of data differ}
\end{figure}

The proof of Algorithm 3 focuses on two main key points: 1) to guarantee that all of $w_{n-\tau}$ to $w_n$ is on one side of the fixed point in the direction of the sample, and 2) to determine the value of the maximum contraction map rate when using this kind of delay SGD. Both of above 2 key points are described in the proof of Lemma 4. 

For the first point, when running the ($n+1$)th update step,  we also need to ensure that the first $n$ update steps satisfy the algorithm. The above requirement means that we should be able to find a range in which the projection of the unique fixed point in the current sample direction addressed. With the processing, the range should shrink. We calculate the range of the fixed point based on the latest iteration information at the beginning of each update step, like figure 5. We only accept the new model parameters that are on the same side of this range as the older model parameters; otherwise, we abandon these new model parameters and use another sample to recalculate new model parameters. The above operation is determined by the point of this range closest to the old model parameters (in Algorithm 3, this point is denoted $Length_{min}$). These processes are described in $Check$ function in Algorithm 4.

For the second point, WP-SGD and Simul Parallel SGD share the same proof frame. In the proof of Simul Parallel SGD, the Lemma 3 in Zinkevich et al.'s work \cite{Zinkevich2010Parallelized} decides the contracting map rate of Simul Parallel SGD. Here, we offer following Lemma 4 for Algorithm 3. Using the proof frame of Simul Parallel SGD with following Lemma 4 instead of Lemma 3 in Zinkevich et al.'s work \cite{Zinkevich2010Parallelized}, we can find the contracting map rate of Algorithm 3 and finish the whole proof.
\begin{figure}[!ht]
	\centering
	\includegraphics[width=9cm,height=3.5cm]{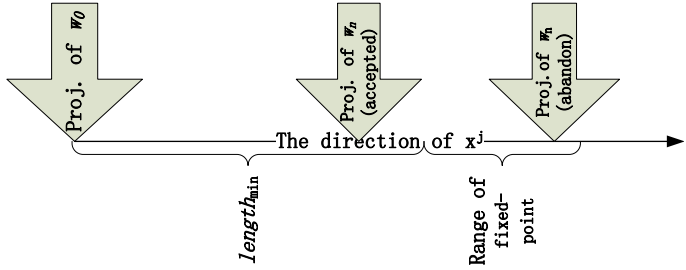}
	\caption{Algorithm 3 only  only accept the new model parameters that are on the same side of this range as the older model parameters ($w_0$ in this figure) }
\end{figure}

The details of Lemma 4's proof are offered in the Appendix.

\textbf{Lemma 4} Let  $c^{*} \geq \left\| \dfrac{{\partial}L(y, \hat y)}{\partial \hat y} \right\|$ be a Lipschitz bound on the loss gradient. Then if $\eta \lambda + \eta \beta^2_{max}{c}^{*} \le {(1 - \eta\lambda)}^{M}$ and Algorithm 3 can consume the whole dataset, the Algorithm 3 is a convergence to the fixed point in $\ell_2$ with Lipschitz constant $1-\lambda\eta$. $\beta_{max}^2$ is defined as $\beta_{max}^2 = \underset{}  {\mathop{\max}}  {\left\|x^i\right\|} ^2$. $M$ is the maximum delay.

If we choose $\eta$ "low enough", gradient descent uniformly becomes a contraction.

However, there exists a $Check$ function in Algorithm 3. $Check$ function suggests that some samples may not be used to trained. Algorithm 3 may be terminated because there is no suitable sample to pass the $Check$ function in dataset. So Algorithm 3 is a theoretical feasibility algorithm instead of a practicable algorithm. 

As we discussed in Section II, the maximum lag the system can tolerate is the maximum number of workers the system can have. When all workers have the same performance, the system will achieve the most efficient working state. In practice, it is very hard to let all nodes in an unbalanced-workload system have the same performance, especially when the clusters consist of different kinds of computing devices. Algorithm 3 is the algorithm designed for this kind of cluster.

\section{Numerical experiments}
We conducted our experiments on a cluster consisting of 10 nodes with a Xeon(R) CPU E5-2660 v2 @ 2.20~GHz, and there was one process on each node.

\subsection{Real-world data}
\begin{table}[!htbp]
	\label{table_real_data}
	\centering
	\begin{tabular}{|l|ccc|}
		\hline
		\diagbox{Iteration}{Obj.}{Method} 	 &SimuParallel	&Weighted SGD &Averaging Directly\\
		\hline
		1	
		&264.925	&264.925  &264.925\\
		100000
		&245.760	    &245.264 &251.062 \\
		200000	
		&229.025	&229.733 &235.224\\
		300000	
		&213.384	&213.612 &222.767\\
		400000	
		&198.697	&198.384 &212.291 \\
		500000	
		&185.311	&185.436 &203.155 \\
		600000	
		&173.795	&173.748 &193.497\\
		700000
		&163.657    &163.702 &184.142\\
		800000
		&154.109    &154.130  &176.307\\
		\hline
	\end{tabular}
	\caption{Using hinge loss training parameters in different parallel SGD algorithms. Obj. is the abbr. of objective function value}
\end{table}

\textbf{Data:}
We performed experiments on KDD Cup 2010 (algebra) \cite{Yu2010Feature}, with labels $y\in \{0,1\}$ and binary, sparse features. The dataset contains 8,407,752 instances for training and 510,302 instances for testing. Those instances have 20,216,830 dimensions. Most instances have about 20--40 features on average.

\textbf{Evaluation measures:}
We chose hinge loss, which is used to train support vector machine (SVM) parameters, as our objective function value. Compared with other loss functions, the contraction map rate of hinge loss is much closer to the contraction map rate of the SGD framework, i.e., ($1-\eta\lambda$).

It is worth noting that our work would be more conspicuous if we use deep learning model like VGG16 \cite{Simonyan2014Very} as our experiment benchmark. But, our paper focuses on the correctness and effectiveness of WP-SGD. There is few work on the mathematical properties of deep learning. If we use deep learning model parameters, we are not sure that the reason for our experiment result is the intricate deep learning network or the effect of WP-SGD.

\textbf {Configurations:}
In the experiment, we set $\lambda = 0.01$, $\eta=0.0001$. And we use $r = 0.99999$. Because the final output is close to the zero vector and we wanted to have more iteration steps, the initial values of all model parameters were set to 4. Testing our algorithms in actual heterogeneous computing environments, such as on a GPU/CPU, is arduous and unnecessary. Since the essence of a heterogeneous computing environment lies in the unbalanced workload of training data consumption of each node, we adopted a software method to simulate the unbalanced-workload environment: In our cluster, the quantity of training data for each of eight nodes (which we call them as the fastest nodes) was five times that for each of the remaining two nodes (which we call them as slow nodes). Then, we studied SVM model parameters and calculated the objective function value on the testing data. As the baseline, we used the output from SimuParallel SGD and the outputs created by using the direct averages of the model parameters. we name latter algorithm as averaging directly.

\textbf{Approach:}
In order to evaluate the convergence speed and hinge loss of the algorithms on an unbalanced-workload system, we used the following procedure: for the configuration, we trained 10 model parameters, each on an independent random permutation of a part of the whole dataset. During training, the model parameters were stored on disk after $k = 100,000 \times i$ updates of the fastest node.

\textbf {Results:}
Table II shows the objective function value of SimuParallel SGD, WP-SGD, and averaging the model parameters directly. In terms of wall clock time, the model parameters obtained on a balanced-workload system, i.e., SimuParallel SGD, clearly outperformed the ones obtained on an unbalanced-workload system. The output of WP-SGD was close to the output on a balanced-workload system. Unsurprisingly, averaging the model parameters directly turned out to be the worst algorithm. The above results are consistent with our Theorem 4, proved in Section III. The convergence speeds of WP-SGD and SimuParallel SGD are the closest. Thus, on an unbalanced-workload system, WP-SGD would obtain a better objective function value. As we can see from the configuration, the time SimuParallel SGD used was five times that used by WP-SGD on the unbalanced-workload system. Therefore, it is feasible and beneficial in practice that  parallelized training model parameters on an unbalanced-workload system with WP-SGD.

\subsection{Analog data}
\begin{figure}[!ht]
	\label{50_font_size_imbalance_svm.eps}
	\centering
	\includegraphics[width=9.5cm,height=4.7cm]{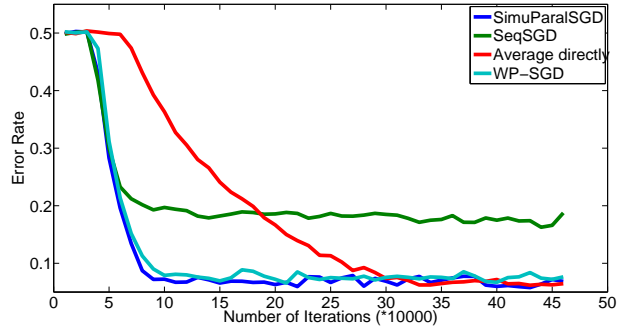}
	\caption{Using SVM model parameters in different SGD algorithms on a cluster with two slow nodes}
\end{figure}

\begin{figure}[!ht]
	\label{50_font_size_svm_some_no_work.eps}
	\centering
	\includegraphics[width=9.5cm, height =4.7cm]{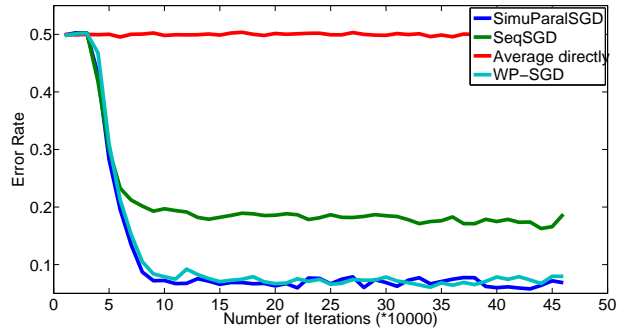}
	\caption{Using SVM model parameters in different SGD algorithms on a cluster with two almost-no-work nodes}
\end{figure}

\textbf {Data:} We performed analog experiments, with labels $y\in \{0,1\}$ and binary, sparse features. The dataset contained 460,000 instances for training and 40,000 instances for testing. Those instances had 100,000 dimensions. Most instances had 5--10 features on average. All of these features, including the position and its value, were generated randomly. Those instances were labeled by $y=\sum\limits_{i=1}^{100000}{{{x}_{i}}\cdot (i\%4)\cdot {{(-1)}^{i}}}$ where $x_i$ is the $i$th feature of sample, with $label = 1$ when $y > 0$ and $label = 0$ otherwise. All instances were normalized to unit length for the experiments.

\textbf {Configurations:}
In the experiment, we set $\lambda = 0.01$, $\eta=0.0001$. And we use $r = 0.99999$. We studied SVM parameters but calculated the prediction error rate on the testing dataset, because the error rate can clearly show the gap between different algorithms and corresponds well to the hinge loss \cite{Bendavid2012Minimizing}. In addition, error rate is the ultimate aim for machine learning. As the baseline, we used sequential SGD, SimuParallel SGD, and averaging the model parameters directly which produce the output by using the direct averages of the model parameters. In this experiment, SimuParallel SGD was used to represent the balanced-workload algorithm. Because we wanted to have more iteration steps, the initial values of the model parameters were set to 4 as the final output is almost the zero vector. In the Figure 6 experiment, the unbalanced-workload setup is the same as in the real-world-data experiment. In the Figure 7 experiment, there existed two nodes that did not work, while eight nodes trained on equal quantities of data.

\textbf{Approach:}
In order to evaluate the convergence speed and error rate of the algorithms on an unbalanced-workload system, we used the following procedure: we trained 10 model parameters, each on an independent, random permutation of a part of the whole dataset. During training, the parameters were stored on disk after $k = 10,000 \times i$ updates of the fastest node. In the Figure 6 experiment, the results show that WP-SGD is still effective in the case where the workload in the system is unbalanced. In Figure 7, the experimental results show that WP-SGD is effective even in the case where the workload in the system is seriously unbalanced.

\textbf {Results:}
Figures 7 and 6 show the error rates of the following algorithms: sequential SGD, SimuParallel SGD (balanced-workload algorithm), WP-SGD, and averaging the model parameters directly. As expected, the balanced-workload algorithm, i.e., SimuParallel SGD, outperformed the sequential SGD and the directly averaged model parameters. The error rate of WP-SGD on the unbalanced-workload system was close to that of SimuParallel SGD on the balanced-workload system. Averaging the model parameters directly was the worst algorithm. In Figure 6, the experimental results show that parallelized training on an unbalanced-workload system benefits from WP-SGD, which converged faster than averaging the models directly.  As we can see from the configuration, the time SimuParallel SGD used is five times that used by WP-SGD on the unbalanced-workload system. In Figure 7, the experimental results show that WP-SGD significantly outperformed averaging the model parameters directly, in terms of convergence speed and algorithm efficiency, when the workload in the system was seriously unbalanced. As we can see from the configuration, SimuParallel SGD cannot work in this environment at all. All of these phenomena correspond well with Theorem 4.

\section{Conclusion}
In this paper, we have proposed WP-SGD, a data-parallel stochastic gradient descent algorithm. WP-SGD inherits the advantages of SimuParallel SGD: little I/O overhead, ideal for MapReduce implementation, superb data locality, and fault tolerance properties. This algorithm also presents strengths in an unbalanced-workload computing environment such as a heterogeneous cluster. We showed in our formula derivation that the upper bound of the objective function value in WP-SGD on an unbalanced-workload system is close to the upper bound of the objective function value in SimuParallel SGD on a balanced-workload system. Our experiments on real-world data showed that the output of WP-SGD was reasonably close to the output on a balanced-workload system. Our experiments on analog data showed that WP-SGD was robust when the workload in the system was seriously unbalanced.
 
For future work, we plan to apply the proposed WP-SGD algorithm for training datasets with higher dimensionality on an actual heterogeneous cluster in a complex network and computing environment. We also plan to design practicable WP-SGD \& traditional parallel SGD algorithms mixed algorithms. What is more, we will take more time in applying WP-SGD on deep learning model parameters which is not well understand on the mathematical properties, like convexity, Lipschitz continuity etc..

\section{Acknowledgment}
This work was supported by the National Natural Science
Foundation of China under Grant No. 61432018, Grant No.
61502450, Grant No. 61521092, and Grant No. 61272136, and by the
National Major Research High Performance Computing
Program of China under Grant No. 2016YFB0200800.

We thank Dr. Fei Teng from ICT, CAS, who gave us several valuable suggestions.

\section{Reference }
\bibliography{mybibfile(1)}

\section*{Appendix}
\textbf{Lemma 2}  Suppose that ${{X}^{1}} \dots {{X}^{k}},B $ are independent distributed random variables over ${\mathbb{R}^{d}}$. Then if $A=\sum\limits_{i=1}^{k}{weigh{{t}_{i}}\cdot {{X}^{i}}}$ and $1=\sum\limits_{i=1}^{k}{weigh{{t}_{i}}}$, it is the case that
\begin{equation}
d({{\mu }_{A}},{{\mu }_{B}})\le \sum\limits_{i=1}^{k}{weigh{{t}_{i}}\cdot d({{\mu }_{{{X}^{i}}}},{{\mu }_{B}})} \notag
\end{equation}

\begin{proof}
	It is well known that if $X^i$ are independent distributed random variables then
	\begin{equation}
	{{\mu }_{A}}=\sum\limits_{i=1}^{k}{weigh{{t}_{i}}\cdot {{\mu }_{{{X}^{i}}}}} \notag
	\end{equation}
	In this proof, we define $vector_{a-b}$ as the vector between the point $a$ and the point $b$.
	Because
	\begin{equation}
	d(a,b)=d(b-vector_{a-b},b)=\left\| {vector_{a-b}} \right\| \notag
	\end{equation}
	it holds that
	\begin{align}
	&d({{\mu }_{A,}}{{\mu }_{B}})=d(\sum\limits_{i=i}^{k}{weigh{{t}_{i}}\cdot {{\mu }_{{{X}^{i}}}},{{\mu }_{B}})} \notag \\
	&=d(\sum\limits_{i=1}^{k}{weigh{{t}_{i}}({{\mu }_{B}}-{vector_{{{\mu }_{B}}-{{\mu }_{{{X}^{i}}}}}})},{{\mu }_{B}}) \notag
	\end{align}
	and 	
	\begin{equation}
	1=\sum\limits_{i=1}^{k}{weigh{{t}_{i}}} \notag
	\end{equation}
	it holds that
	\begin{align}
	&d(\sum\limits_{i=1}^{k}{weigh{{t}_{i}}({{\eta }_{B}}-{{m}_{{{\eta }_{B}}-{{\eta }_{{{X}^{i}}}}}})},{{\eta }_{B}}) \notag \\
	&=d({{\eta }_{B}}-\sum\limits_{i=1}^{k}{weigh{{t}_{i}}\cdot {{m}_{{{\eta }_{B}}-{{\eta }_{{{X}^{i}}}}}}},{{\eta }_{B}}) \notag \\
	&=\left\| \sum\limits_{i=1}^{k}{weigh{{t}_{i}}\cdot {{m}_{{{\eta }_{B}}-{{\eta }_{{{X}^{i}}}}}}} \right\|\le \sum\limits_{i=1}^{k}{weigh{{t}_{i}}}\cdot \left\| {{m}_{{{\eta }_{B}}-{{\eta }_{{{X}^{i}}}}}} \right\| \notag \\
	&=\sum\limits_{i=1}^{k}{weigh{{t}_{i}}\cdot d({{\eta }_{{{X}^{i}}}},{{\eta }_{B}})}\notag
	\end{align}
\end{proof}

\textbf{Corollary 1} The fastest node consumes $t$ data samples, $D_{\eta }^{t-{{T}_{i}}}$ is the distribution of model parameters updated after $t-{{T}_{i}}$ iterations in node $i$, and $D_{\eta }^{\#,t}$ is the distribution of the stochastic gradient descent update in WP-SGD.
\begin{equation}
d({{\eta }_{D_{\eta }^{\#,t}}},{{\eta }_{D_{\eta }^{*}}})\le \frac{Gk{{(1-\eta \lambda )}^{t}}}{\lambda \sum\limits_{i=1}^{k}{{{(1-\eta \lambda )}^{{{T}_{i}}}}}} \notag
\end{equation}

\begin{proof}
	Suppose ${{W}^{\#,t}}$ is the output of the algorithm, and ${{W}^{i,t-{{T}_{i}}}}$ is the output of each node. Then
	\begin{equation}
	\notag	{{W}^{\#,t}}=\sum\limits_{i=1}^{k}{weigh{{t}_{i}}\cdot {{W}^{i,t-{{T}_{i}}}}}
	\end{equation}
	and therefore
	\begin{equation}
	\notag	d({{\eta }_{D_{\eta }^{t-{{T}_{i}}}}},{{\eta }_{D_{\eta }^{*}}})\le \frac{G}{\lambda }{{(1-\eta \lambda )}^{t-{{T}_{i}}}}
	\end{equation}
	Thus, using Lemma 2,
	\begin{equation}
	\notag	d({{\eta }_{D_{\eta }^{\#,t}}},{{\eta }_{D_{\eta }^{*}}})\le \frac{G}{\lambda }\sum\limits_{i=1}^{k}{weigh{{t}_{i}}\cdot {{(1-\eta \lambda )}^{t-{{T}_{i}}}}}
	\end{equation}
	Combining the above with the definition of
	\begin{equation}
	\notag	weight_{1-\lambda \eta,i}^{{}}={{(1-\lambda \eta )}^{{{T}_{i}}}}/\sum\limits_{j=1}^{k}{{{(1-\lambda \eta )}^{{{T}_{j}}}}}
	\end{equation}
	we have
	\begin{equation}
	\notag	d({{\eta }_{D_{\eta }^{\#,t}}},{{\eta }_{D_{\eta }^{*}}})\le \frac{Gk{{(1-\eta \lambda )}^{t}}}{\lambda (\sum\limits_{i=1}^{k}{{{(1-\eta \lambda )}^{{{T}_{i}}}}}}
	\end{equation}
	
\end{proof}

\textbf{Lemma 3}  ${{W}^{i,t-{{T}_{i}}}}$ is the output of node $i$. Then, if
\begin{equation}
\notag	{{W}^{\#,t}}=\sum\limits_{i=1}^{k}{weigh{{t}_{1-\eta \lambda,i}}\cdot {{W}^{i,t-{{T}_{i}}}}}
\end{equation}
then the distribution of ${W}^{\#,t}$ is $D^{\#,t}_{\eta}$. It is the case that
\begin{equation}
\notag	\sigma _{{{D_\eta}^{\#,t}}}\le \frac{\sqrt{k}}{(\sum\limits_{i=1}^{k}{{{(1-\eta \lambda )}^{{{T}_{i}}}})}}(\frac{2G\sqrt{\eta }}{\sqrt{\lambda }}+\frac{G}{\lambda }{{(1-\eta \lambda )}^{t}})
\end{equation}

\begin{proof}
	\begin{equation}
	\notag	\sigma _{{{W}^{\#,t}}}^{2}=\sum\limits_{i=1}^{k}{weight_{1-\lambda \eta,i}^{2}\cdot \sigma _{{{W}^{i,t-{{T}_{i}}}}}^{2}}
	\end{equation}
	Combining this with Theorem 3, we obtain
	\begin{align}
	& weigh{{t}_{1-\eta \lambda,i}}\cdot {{\sigma }_{{{W}^{i,t-{{T}_{i}}}}}} \notag \\
	&\le weigh{{t}_{1-\eta \lambda,i}}\cdot (\frac{2\sqrt{\eta }G}{\sqrt{\lambda }}+\frac{G}{\lambda }{{(1-\eta \lambda )}^{t-{{T}_{i}}}}) \notag \\
	& =\frac{1}{\sum\limits_{j=1}^{k}{{{(1-\eta \lambda )}^{j}}}}(\frac{2G\sqrt{\eta }}{\sqrt{\lambda }}\cdot {{(1-\eta \lambda )}^{{{T}_{i}}}}+\frac{G}{\lambda }{{(1-\eta \lambda )}^{t}}) \notag \\
	& \le \frac{1}{\sum\limits_{j=1}^{k}{{{(1-\eta \lambda )}^{j}}}}(\frac{2G\sqrt{\eta }}{\sqrt{\lambda }}+\frac{G}{\lambda }{{(1-\eta \lambda )}^{t}}) \notag
	\end{align}
	Thus,
	\begin{equation}
	\sigma _{{{W}^{\#,t}}}^{2}\le \frac{k}{(\sum\limits_{i=1}^{k}{{{(1-\eta \lambda )}^{{{T}_{i}}}}{{)}^{2}}}}{{(\frac{2G\sqrt{\eta }}{\sqrt{\lambda }}+\frac{G}{\lambda }{{(1-\eta \lambda )}^{t}})}^{2}} \notag
	\end{equation}
\end{proof}

\textbf{Theorem 4} Given a cost function $c$ such that ${{\left\| c \right\|}_{\mathrm{Lip}}}$ and ${{\left\| \nabla c \right\|}_{\mathrm{Lip}}}$ are bounded, the bound of WP-SGD is
\begin{align}
& {{E}_{w\in D}}[c(w)]-\underset{w}{\mathop{\min }}\,c(w) \notag \\
&\le ((\frac{Gk{{(1-\eta \lambda )}^{t}}}{\lambda(\sum\limits_{j=1}^{k}{{{(1-\eta \lambda )}^{{{T}_{j}}}}})}+\frac{\sqrt{k}}{\sum\limits_{j=1}^{k}{{{(1-\eta\lambda)}^{{{T}_{j}}}}}}(\frac{2G\sqrt{\eta }}{\sqrt{\lambda }} \notag
\\
& +\frac{G}{\lambda }{{(1-\eta \lambda )}^{t}}))\sqrt{2{{\left\|\nabla c \right\|}_{\mathrm{Lip}}}}+  \sqrt{c(v)-\underset{w}{\mathop{\min }}\,c(w)}{{)}^{2}} \notag
\end{align}
\begin{proof}
	Theorem 1 offers the upper bound of the fixed point and the minimum of the objective function which is controlled by relative standard deviation. Lemma 1 is the upper bound of relative standard deviation which is controlled by mean and standard deviation. Lemma 3 and Corollary 1 are the upper bound of the mean and standard deviation controlled by the number of iterations. Combining all of them, we easily obtain Theorem 4.
\end{proof}

\textbf{Corollary 2}  For WP-SGD, when
\begin{align}
&\dfrac{\sum\limits_{i=1}^{k}{{{(1-\eta \lambda )}^{{{T}_{i}}}}}- \sqrt{k}}{k-\sum\limits_{i=1}^{k}{{{(1-\eta \lambda )}^{{{T}_{i}}}}}} \notag \\
&> \dfrac{(1-\lambda\eta)^t Wasserstein_1({{D_{\eta }^{\#,1}}},{{D_{\eta }^{*}}}) }{(1-\lambda\eta)^t \cdot Wasserstein_2({{D_{\eta }^{\#,1}}},{{D_{\eta }^{*}}}) + \sigma_{D_\eta^*}} \notag
\end{align}
the upper bound of the objective function value of WP-SGD is closer to the minimum than is the upper bound of the objective function value of sequential SGD on the fastest nodes.
\begin{proof}
	We use the upper bound of the relative standard deviations from Lemma 30 in Zinkevich et al.'s work \cite{Zinkevich2010Parallelized}. The upper bound of the objective function value of SGD is positively correlated with the relative standard deviations. Thus, when the upper bound of the relative standard deviations of WP-SGD is less than that of sequential SGD, we obtain this corollary.
\end{proof}

\textbf{Corollary 3} For WP-SGD, on a dataset having a large standard deviation and a large standard deviation of the fixed point, when
\begin{equation}
\notag	2\sum\limits_{i=1}^{k}{{{(1-\eta \lambda )}^{{{T}_{i}}}}}>\sqrt{k}+k
\end{equation}
the upper bound of the objective function value of WP-SGD is closer to the minimum than is the upper bound of the objective function value of sequential SGD on the fastest nodes.

\begin{proof}
	Notice that ${{E}_{w\in D}}[c(w)]-\underset{w}{\mathop{\min }}\,c(w)$ decreases as the first part of Theorem 4 decreases. The first part of Theorem 4 which also can be written in the following way.
	\begin{equation}
	\frac{G}{\lambda }(\frac{k+\sqrt{k}}{\sum\limits_{j=1}^{k}{{{(1-\eta \lambda )}^{{{T}_{i}}}}}}{{(1-\eta \lambda )}^{{{T}_{i}}}}+\frac{2\sqrt{k}}{\sum\limits_{j=1}^{k}{{{(1-\eta \lambda )}^{{{T}_{i}}}}}}\sqrt{\eta \lambda }) \notag
	\end{equation}
	In addition, the sequential algorithms are a special case in WP-SGD when $k = 1$. Thus, if WP-SGD is better than the sequential algorithm, the first part of Theorem 4 must be less than
	\begin{equation}
	\frac{G{{(1-\eta \lambda )}^{t}}}{\lambda }+(\frac{2G\sqrt{\eta }}{\sqrt{\lambda }}+\frac{G}{\lambda }{{(1-\eta \lambda )}^{t}}) \notag
	\end{equation}
	which can be written as
	\begin{equation}
	\frac{G}{\lambda }(2{{(1-\eta \lambda )}^{t}}+2\sqrt{\eta \lambda }) \notag
	\end{equation}
	It is apparent that if following inequalities hold, we obtain the result.
	\begin{equation}
	\notag		\frac{k+\sqrt{k}}{\sum\limits_{i=1}^{k}{{{(1-\eta \lambda )}^{{{T}_{i}}}}}}\le 2
	\end{equation}
	and
	\begin{equation}
	\notag		\frac{\sqrt{k}}{\sum\limits_{i=1}^{k}{{{(1-\eta \lambda )}^{{{T}_{i}}}}}}\le 1
	\end{equation}
	which means
	\begin{equation}
	\notag		2\sum\limits_{i=1}^{k}{{{(1-\eta \lambda )}^{{{T}_{i}}}}}>\sqrt{k}+k
	\end{equation}
\end{proof}

\textbf{Deduction 1} Given a cost function $c$ such that ${{\left\| c \right\|}_{\mathrm{Lip}}}$ and ${{\left\| \nabla c \right\|}_{\mathrm{Lip}}}$ are bounded, we average parameters every $span$ iterations for the fastest node in SimuParallel SGD. Then, the bound of the algorithm is
\begin{align}
&{{E}_{w\in D_{\eta }^{T,k}}}[c(w)]-\underset{w}{\mathop{\min }}\,c(w)\le \notag\\
& ((\frac{G{{(1-\eta \lambda )}^{t}}}{\lambda }+\frac{1}{{{(\sqrt{k})}^{t/span}}}(\frac{2G\sqrt{\eta }}{\sqrt{\lambda }} \notag \\
&+\frac{G}{\lambda }{{(1-\eta \lambda )}^{t}}))\sqrt{2{{\left\| \nabla c \right\|}_{\mathrm{Lip}}}} \notag\\
\notag	&+\sqrt{c(w){{|}_{w\in D_{\eta }^{D,k}}}-\underset{w}{\mathop{\min }}\,c(w)}{{)}^{2}}
\end{align}
\begin{proof}
	Every averaging operation reduces the variance by $1/\sqrt{k}$, and every iteration step reduces the Euclidean distance and part of the variance by $(1-\eta \lambda )$. Thus, we obtain Deduction 1.
\end{proof}
\textbf{Deduction 2} Given a cost function $c$ such that ${{\left\| c \right\|}_{\mathrm{Lip}}}$ and ${{\left\| \nabla c \right\|}_{\mathrm{Lip}}}$ are bounded, we average parameters every $span$ iterations for the fastest node in WP-SGD. Then, the bound of the algorithm is
\begin{align}
& {{E}_{w\in D}}[c(w)]-\underset{w}{\mathop{\min }}\,c(w) \notag \\
&\le ((\frac{G{{(1-\eta \lambda )}^{t}}}{\lambda }\cdot {{(\frac{k}{(\sum\limits_{j=1}^{k}{{{(1-\eta \lambda )}^{{{T}_{j}}}}})})}^{t/span}}+ \notag \\
& {{(\frac{\sqrt{k}}{\sum\limits_{j=1}^{k}{{{(1-\eta \lambda )}^{{{T}_{j}}}}}})}^{t/span}}(\frac{2G\sqrt{\eta }}{\sqrt{\lambda }}+\frac{G}{\lambda }{{(1-\eta \lambda )}^{t}}))\sqrt{2{{\left\| \nabla c \right\|}_{\mathrm{Lip}}}} \notag \\
& +\sqrt{c(v)-\underset{w}{\mathop{\min }}\,c(w)}{{)}^{2}} \notag
\end{align}
\begin{proof}
	Every averaging operation reduces the variance by $\frac{\sqrt{k}}{\sum\limits_{j=1}^{k}{{{(1-\eta \lambda )}^{{{T}_{j}}}}}}$. Every iteration steps reduce the Euclidean distance and part of the variance by $(1-\eta \lambda )$. We obtain the final deduction.
\end{proof}

\textbf{Lemma 4} Let  $c^{*} \geq \left\| \dfrac{{\partial}L(y, \hat y)}{\partial \hat y} \right\|$ be a Lipschitz bound on the loss gradient. Then if $\eta \lambda + \eta {\beta_{max}^2}{c}^{*} \le {(1 - \eta\lambda)}^{M}$ and Algorithm 3 can consume the whole dataset, the Algorithm 3 is a convergence to the fixed point in $\ell_2$ with Lipschitz constant $1-\lambda\eta$. Where $\beta_{max}^2$ is defined as $\beta_{max}^2 = \underset{}  {\mathop{\max}}  {\left\|x^i\right\|} ^2$.
\begin{proof}
	Firstly, by gathering terms, we obtain
	\begin{equation}
	{w}_{n+1} = {w}_{n} - \eta \lambda {w}_{n-\tau} - \eta {x}^{j}\dfrac{\partial}{\partial \hat y}L({{y}^{j}},{\hat y})\left.{}{}\right|_{{w}_{n-\tau}{x}^{j}} \notag
	\end{equation}
	Define $u:R\mapsto \mathbb{R}$ to be equal to $u(z) = \dfrac{\partial}{\partial Z}L({y}^{i},z)$. Because $L(y^i,\hat y)$ is convex in $\hat y$, $u(z)$ is increasing, and $u(z)$ is Lipschitz continuous with constant $c^{*}$.\
	\begin{equation}
	{w}_{n+1} = {w}_{n} - \eta \lambda {w}_{n-\tau} - \eta {x}^{j}u(w_{n-\tau}x^{j}) \notag
	\end{equation}
	We break down $w$ into $w_{\perp}$ and $w_{\parallel}$, and $w_{\parallel}$ is parallel with simple $x^{j}$, where $w = w_{\perp} + w_{\parallel}$. Thus,
	\begin{equation}
	{w}_{n+1 \parallel} = {w}_{n\parallel} - \eta \lambda {w}_{n-\tau\parallel} - \eta {x}^{j}u(w_{n-\tau\parallel}x^{j}) \notag
	\end{equation}
	\begin{equation}
	{w}_{n+1 \perp} = {w}_{n\perp} - \eta \lambda {w}_{n-\tau\perp} \notag
	\end{equation}
	Finally, note that $d(w,v) =\sqrt{d^2(w_\parallel,v_\parallel)+d^2(w_\perp,v_\perp)}$
	For the vertical dimension, we know that
	\begin{align}
	&\dfrac{{w}_{n+1 \perp}}{{w}_{n\perp}} = 1 - \eta \lambda \dfrac {{w}_{n-\tau\perp}}{{w}_{n\perp}} \notag \\
	&=1 - \eta \lambda \dfrac {{w}_{n-1\perp}}{{w}_{n\perp}}*\dfrac {{w}_{n-2\perp}}{{w}_{n-1\perp}}* \cdots *\dfrac {{w}_{n-\tau\perp}}{{w}_{n-\tau+1\perp}}<1-\eta\lambda \notag
	\end{align}
	For the vertical dimension, it is a homogeneous linear recurrence relation. When $n$ is large enough, $\dfrac {{w}_{n-1\perp}}{{w}_{n\perp}} >1$. When the $n$ is not large enough, above requirement is guaranteed by $Check$ function in Algorithm 4.
	
	Now, we focus on the dimension parallel to $x^j$. We define $\alpha(w_n) = x^j \cdot w_n$ (in Algorithm 3, it is $Length_{n}$, and it is the projection of $w_n$ on $x^j$), so we can know that
	\begin{equation}
	\notag		\alpha(w_{n+1}) = \alpha(w_n)-\eta\lambda\alpha(w_{n-\tau})-\eta u(\alpha(w_{n-\tau}))\beta^2
	\end{equation}
	In the Hogwild! algorithm, we know this kind of delay SGD must have a fixed point, and we denote this fixed point by $v$:
	\begin{equation}
	\notag		d(w_\parallel,v_\parallel) = \dfrac{1}{\beta}\vert \alpha(w) - \alpha(v) \vert
	\end{equation}
	\begin{equation}
	\notag		d(w_{n+1\parallel},v_\parallel) = \dfrac{1}{\beta}\vert (w_n - \eta(\lambda \alpha(w_{n-\tau}))) - (v - \eta(\lambda \alpha(v)))\vert
	\end{equation}
	Without loss of generality, assume that $\alpha(w_i)\geq \alpha(v)$ with all $i < n+1$ is true. Since $\alpha(w_n)\geq \alpha(v)$, $u(\alpha(w_n) \geq u(\alpha(v))$. By Lipschitz continuity,
	\begin{equation}
	u(\alpha(w_n))-u(\alpha(v))\leq c^*(\alpha(w_n)-\alpha(v)) \notag
	\end{equation}
	Here, we define
	\begin{equation}
	dis_n = \alpha(w_n) - \alpha(v) \notag \notag
	\end{equation}
	Because of the assumption, we know that $dis_n \geq 0$, and at the beginning, $w_0 = 0$, which means that $length_0 = 0$. The following operation is to provide a rough idea of the range of $v$. What we care about is the range of $v$ closest to $w_0$, which we denote by $length_{min}$. $M$ is the maximum delay.
	
	A rearranging of the terms yields
	\begin{equation}
	dis_{n+1} = \vert dis_n -\eta\lambda dis_{n-\tau}-\eta\beta^2(u(\alpha(w_{n-\tau}))-u(\alpha(v)))\vert \notag
	\end{equation}
	To be able to eliminate the absolute value brackets, we need the terms in the absolute value brackets to be positive. Because $u(\alpha(w_n))-u(\alpha(v))\leq c^*(\alpha(w_n)-\alpha(v))$
	if
	\begin{equation}
	dis_n -\eta\lambda dis_{n-\tau}-\eta\beta^2(u(\alpha(w_{n-\tau}))-u(\alpha(v)))>0 \notag
	\end{equation}
	it follows that
	\begin{equation}
	\dfrac{dis_n}{dis_{n-\tau}}>\eta\lambda+\eta\beta^2c^* \notag
	\end{equation}
	To satisfy the above terms, we require that
	\begin{equation}
	\dfrac{dis_n}{dis_{n-1}}>\sqrt[\tau]{\eta\lambda+\eta\beta^2c^*} \notag
	\end{equation}
	Above requirement is guaranteed by $Check$ function in Algorithm 4.
	
	Above requirement can be rewritten as
	\begin{equation}
	Length_{min}=\dfrac{Length_{t-1} - rate * Length_{t-2}}{ 1 - rate } \notag
	\end{equation}

	Note that on the assumption, $\alpha(w_n))>\alpha(v)$, and so
	\begin{equation}
	dis_{n+1}\leq dis_n - \eta\lambda dis_{n-\tau} \notag
	\end{equation}
	It is apparent that $dis_n$ is a non-increasing serier, which means that
	\begin{equation}
	\dfrac{dis_{n+1}}{dis_n}<1-\eta\lambda \notag
	\end{equation}
	It is apparent that $\eta$ should satisfy
	\begin{equation}
	\eta\lambda+\eta\beta^2 c^* <\dfrac{dis_n}{dis_{n-\tau}}<(1-\eta\lambda)^\tau \notag
	\end{equation}
	and from the whole dateset aspect, and $\tau$ reach the maximum, $\eta$ should satisfy
	\begin{equation}
	\eta\lambda+\eta\beta_{max}^2 c^* <\dfrac{dis_n}{dis_{n-\tau}}<(1-\eta\lambda)^M \notag
	\end{equation}	
	and this then implies that
	\begin{align}
	&d(w_{n+1\parallel},v)=\dfrac{1}{\beta}(\alpha(w_n+1)-\alpha(v))\notag \\
	 &\leq(1-\eta\lambda)\dfrac{1}{\beta}(\alpha(w_n)-\alpha(v))=(1-\eta\lambda)d(w_{n\parallel},v_\parallel)\notag
	\end{align}
\end{proof}


\end{document}